\documentclass[11pt]{article}
\usepackage{arxiv}
\usepackage{graphicx}
\usepackage[linesnumbered]{algorithm2e}

\usepackage{amsmath}
\usepackage{amssymb}
\usepackage{mathtools}
\usepackage{amsthm}
\usepackage{bm}
\usepackage{multirow}
\usepackage{subfigure}
\usepackage{booktabs}
\usepackage{adjustbox}
\usepackage[dvipsnames]{xcolor}
\usepackage{cancel}
\usepackage{soul}
\usepackage[hidelinks]{hyperref}
\setstcolor{black}

\theoremstyle{plain}
\newtheorem{theorem}{Theorem}

\theoremstyle{definition}

\newtheorem{assumption}[theorem]{Assumption}
\theoremstyle{remark}

\newcommand{\bs}[1]{\boldsymbol{#1}}

\title{StoCFL: A Stochastically Clustered Federated Learning Framework for Non-IID Data with Dynamic Client Participation}

\author{%
Dun Zeng$^{1,3}$ \quad Xiangjing Hu$^4$ \quad Shiyu Liu$^{1,3}$ \\
Yue Yu$^3$ \quad Qifan Wang$^6$ \quad Zenglin Xu$^{2,3,5}$\thanks{Corresponding author: \texttt{zenglin@gmail.com}} \\[0.6em]
\small $^1$University of Electronic Science and Technology of China, Chengdu, China \\
\small $^2$Fudan University, Shanghai, China \\
\small $^3$Peng Cheng Laboratory, Shenzhen, China \\
\small $^4$Harbin Institute of Technology, Shenzhen, China \\
\small $^5$Shanghai Academy of AI for Science, Shanghai, China \\
\small $^6$Meta AI, Menlo Park, California, USA \\[0.4em]
\footnotesize \texttt{zengdun@std.uestc.edu.cn; xiangjinghu@stu.hit.edu.cn; shyu.liu@foxmail.com} \\
\footnotesize \texttt{yuyue@nudt.edu.cn; wqfcr618@gmail.com}
}
\date{}

\begin{document}
\maketitle

{\centering\small Author preprint. Published version: \emph{Neural Networks} 187 (2025), 107278.\par
\url{https://doi.org/10.1016/j.neunet.2025.107278}\par}
\vspace{0.5em}

\begin{abstract}
Federated learning is a distributed learning framework that takes full advantage of private data samples kept on edge devices. In real-world federated learning systems, these data samples are often decentralized and Non-Independently Identically Distributed (Non-IID), causing divergence and performance degradation in the federated learning process. As a new solution, clustered federated learning groups federated clients with similar data distributions to impair the Non-IID effects and train a better model for every cluster. However, existing CFL algorithms are ineffective because they lack an information-sharing mechanism across clusters resulting in low data efficiency and model performance. Meanwhile, their performance is highly subjected to ideal client clustering results which are practically unavailable. This paper proposes StoCFL, a novel clustered federated learning framework for generic Non-IID issues. In detail, StoCFL implements a flexible CFL framework that supports an arbitrary proportion of client participation and newly joined clients for a varying FL system, while maintaining a great improvement in model performance. The intensive experiments are conducted by using four basic Non-IID settings and a real-world dataset. The results show that StoCFL could obtain promising cluster results even when the number of clusters is unknown. Based on the client clustering results, models trained with StoCFL outperform baseline approaches in a variety of scenarios.
\end{abstract}


\section{Introduction}
Federated Learning (FL)~\citep{mcmahan2017communication, yang2019federated, cai2023stereo, hoang2024effective} enables smart devices or institutions to collaboratively conduct machine learning tasks without compromising privacy regulations~\citep{zhang2023survey, zhang2024survey}. This approach maximizes the utilization of data collected on mobile phones and personal computers within the framework. Moreover, these data samples often encompass habits, preferences, and even geographic information~\citep{zeng-etal-2024-diversified, DBLP:journals/ftml/KairouzMABBBBCC21, lms, pleasing}. Consequently, the distribution of data samples among devices in this decentralized system can vary significantly, which is also termed as data Non-Independently Identically Distributed (Non-IID).
Existing federated learning~\citep{mcmahan2017communication, zhao2018federated, DBLP:conf/mlsys/LiSZSTS20, LiHYKLXN22, li2023revisiting, zeng2023power} has theoretically demonstrated that Non-IID data distribution across clients leads to substantial gradient variance. 
Consequently, this variance significantly increases the number of communication rounds required to achieve model convergence. Empirical findings further indicate that Non-IID data severely degrades the generalization performance of FL models. Therefore, understanding and addressing gradient relationships across FL clients is crucial for enhancing FL performance in the presence of Non-IID challenges.

In response, \citet{sattler2020clustered} introduced a novel approach termed Clustered Federated Learning (CFL), and its follow-up works~\citep{sattler2020clustered, sattler2020byzantine, ghosh2020efficient, briggs2020federated, duan2021flexible, xie2020multi} have established as a promising solution for mitigating Non-IID issues in FL.
CFL is based on a key intuition: \textbf{although data distributions across FL clients are heterogeneous due to factors such as habits, preferences, and geographic differences, the local datasets of certain clients can be similar enough to collaboratively enhance FL performance without exacerbating Non-IID issues.} Formally, CFL operates under an assumption that enables effective clustering of clients with comparable data distributions.
\begin{assumption}[Clustered Federated Learning~\citep{sattler2020clustered}] \label{asp:cfl}
There exists a partitioning $\mathcal{C} = \{c_1,\dots,c_K\}, \bigcup_{k=1}^K c_k = \{1,\dots,N\}$ ($N \geq K \geq 2$) of the client population, such that every subset of clients $c_k \in \mathcal{C}$ satisfies the distribution learning assumption (i.e., the data distribution among these clients is similar). $N$ is the number of clients and $K$ is the number of clusters.
\end{assumption}

\begin{figure}[t]
\centering
\subfigure[Stochastic Federated Client Clustering]{
\includegraphics[width=0.3\linewidth]{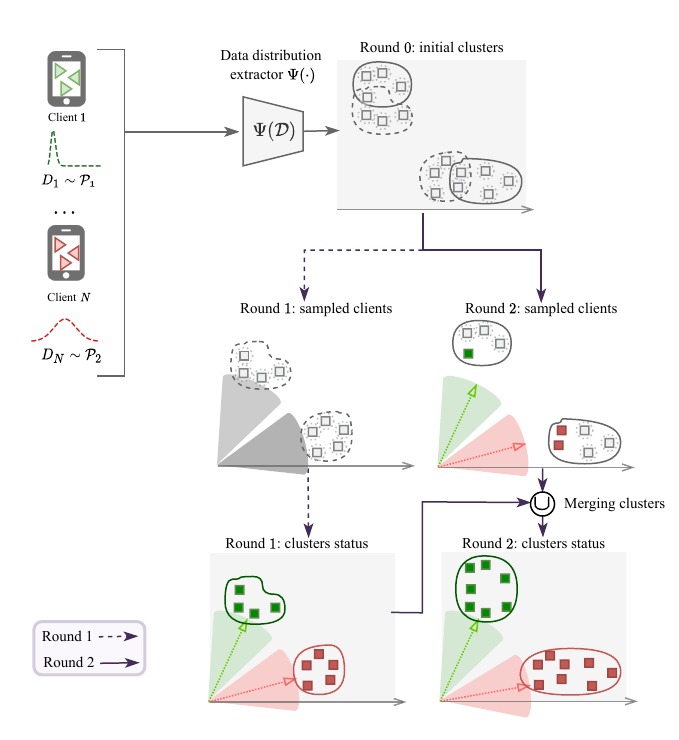}\label{fig:scc}}
\subfigure[Bi-level Clustered Federated Learning]{
\includegraphics[width=0.6\linewidth]{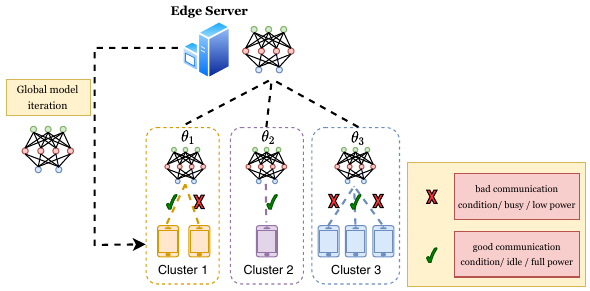}\label{fig:bcfl}
}
\caption{StoCFL consists of two vital components: stochastic client clustering and bi-level clustered federated learning. In Figure \ref{fig:scc}, the distribution extractor $\Psi(\cdot)$ draws local data distribution representation. Then, the server randomly samples a subset of clients for each round and merges clients into clusters based on cosine similarity. Compared with conventional CFL algorithms (in dashed boxes), Figure \ref{fig:bcfl} illustrates an example with 3 clusters that StoCFL enables cluster models to improve each other via a global model.}
\label{fig:cluster}
\end{figure}

When client clustering satisfies the underlying assumptions, empirical results~\citep{sattler2020clustered} demonstrate that the vanilla CFL framework significantly improves FL performance in addressing Non-IID issues. Furthermore, theoretical analyses~\citep{ma2022convergence, ghosh2020efficient} confirm that clustering techniques effectively reduce gradient variance within clusters, facilitating faster convergence. However, vanilla CFL frameworks~\citep{sattler2020clustered, ghosh2020efficient, xie2020multi} train separate FL models on intra-cluster data samples, relying solely on the corresponding clients' data within each cluster. This approach presents a potential risk: clusters with limited data may struggle to achieve the precision needed for effective modeling. Given that model generalization performance is closely tied to the quantity of available data samples~\citep{hardt2016train, sun2024stability, zeng2024understanding}, vanilla CFL suffers from low data efficiency and suboptimal generalization performance.

Existing client clustering approaches often overlook practical constraints, such as communication and computational limitations, which are critical in real-world FL applications (Figure~\ref{fig:bcfl}). For example, prior CFL methods~\citep{sattler2020clustered, stallmann2022towards, duan2021flexible} assume that all clients participate in the FL process to enable client clustering algorithms. However, in realistic FL scenarios, particularly with a large number of edge devices, clients are not always accessible to the server. Additionally, some approaches~\citep{ghosh2020efficient, wang2022federated, mansour2020three, xie2020multi} rely on prior knowledge of the number of clusters, which is often unknown to the server in practice. As the performance of vanilla CFL frameworks is highly dependent on accurate client clustering, misestimating the number of clusters can significantly degrade their FL model performance (see Section~\ref{sec:cfl_exp}).
Given these limitations of existing CFL algorithms, there is a pressing need to explore a flexible and practical CFL framework to enhance data efficiency and model performance under real-world FL constraints.

\begin{figure}[t]
\centering
\subfigure[Vanilla FL]{
\includegraphics[width=0.48\linewidth]{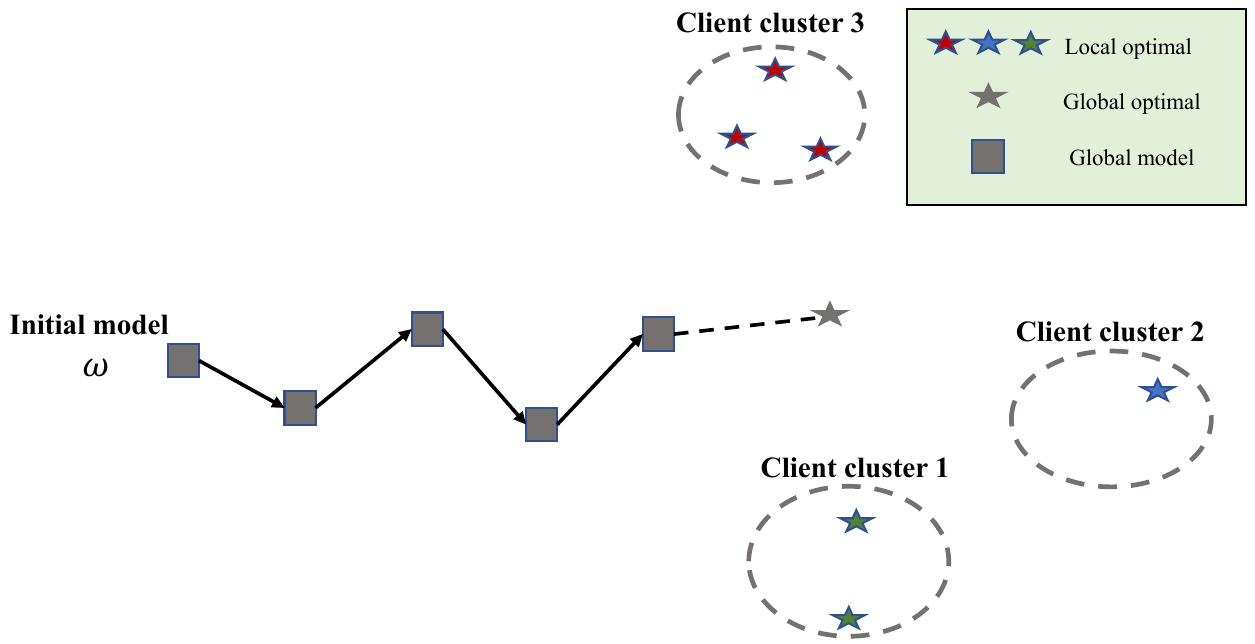}\label{fig:vanillafl}}
\subfigure[Vanilla CFL]{
\includegraphics[width=0.48\linewidth]{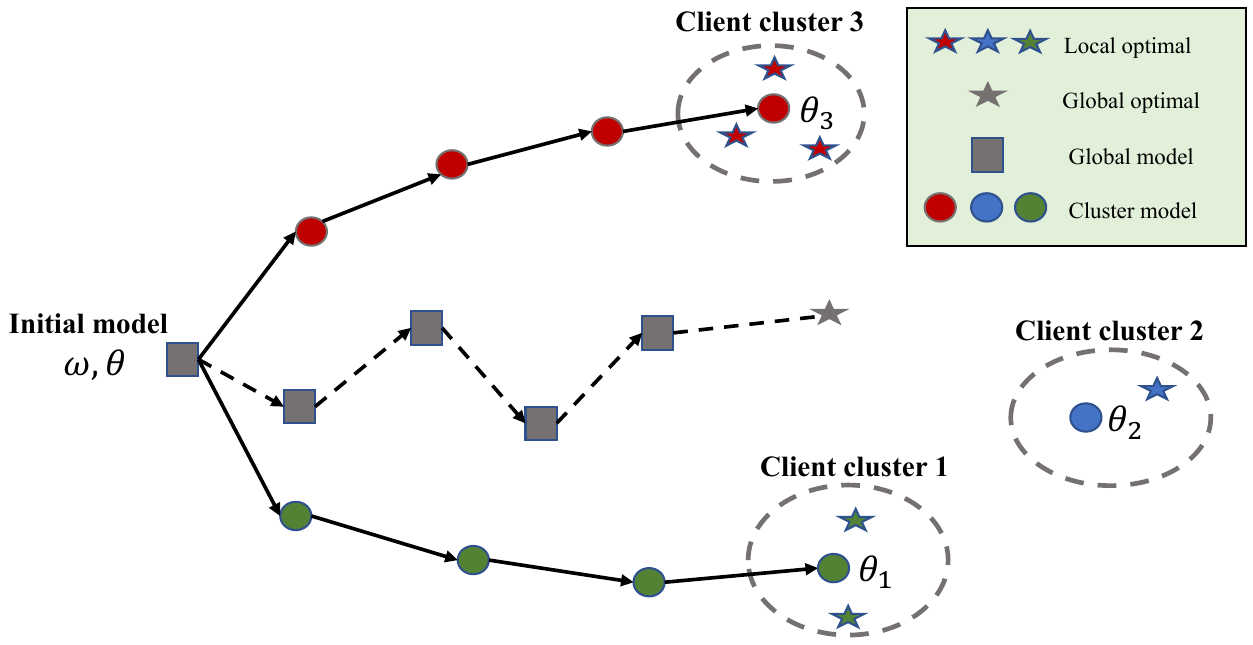}\label{fig:vanillacfl}}
\vspace{2mm}
\subfigure[StoCFL]{
\includegraphics[width=0.48\linewidth]{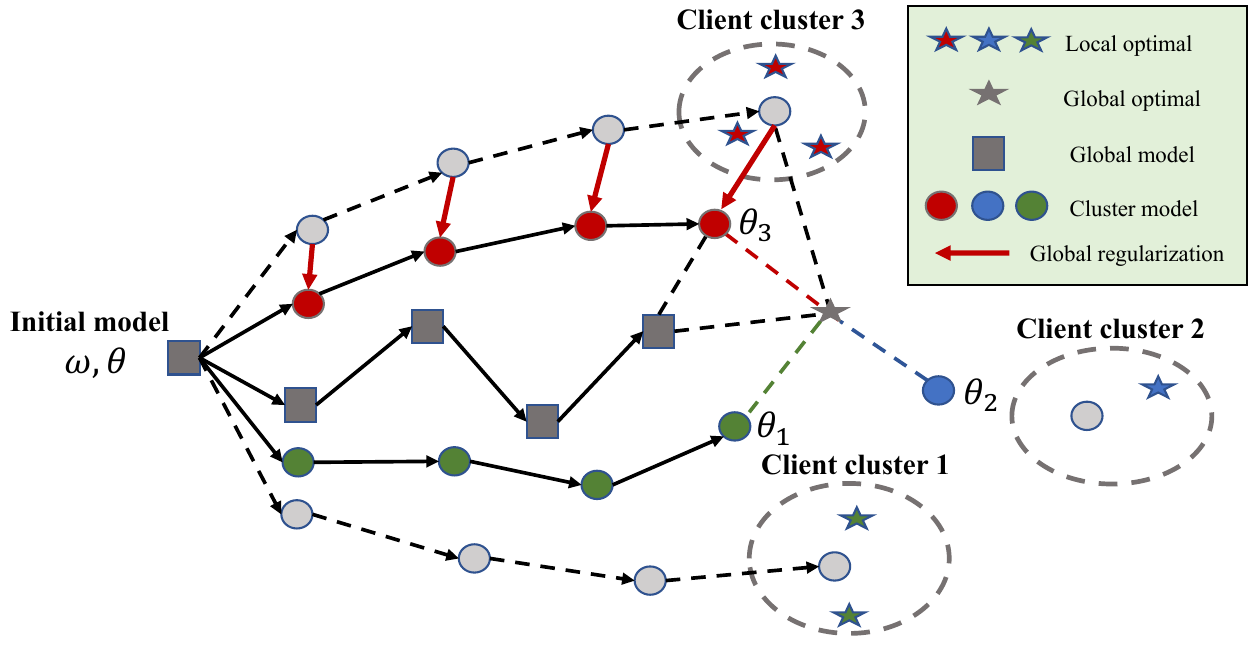}\label{fig:stocfl}
}
\caption{Illustration of StoCFL in the same setting as Figure~\ref{fig:bcfl} (6 clients forming 3 implicit clusters). Figure 2(a) demonstrates that vanilla FL converges slowly to a suboptimal solution due to the heterogeneity of local optima. Figure 2(b) highlights that intra-cluster models in vanilla CFL converge more rapidly, benefiting from reduced data heterogeneity among clients within each cluster. Figure 2(c) showcases how StoCFL leverages global model regularization to effectively balance cluster-level personalization and global model generalization, resulting in superior performance. For clarity, we omit the regularization specifics for $\theta_1$, $\theta_2$, and the iteration trajectory of $\theta_2$.}
\label{fig:stocfl-highlevel}
\end{figure}

To this end, we propose a novel CFL framework termed StoCFL as shown in Figure~\ref{fig:cluster}. 
The proposed StoCFL framework integrates two key components: \textbf{stochastic federated client clustering (SFCC)} and \textbf{Bi-level clustered federated learning (Bi-CFL)} modules.
In detail, SFCC generates a representation vector of the local data distribution for each client and evaluates the similarity between clients using cosine similarity. This module iteratively clusters clients based on their similarity, even when only a subset of clients participates in each FL communication round, as illustrated in Figure~\ref{fig:scc}.
Then, Bi-CFL uses an independently-trained global FL model to regularize the cluster models' training as shown in Figure~\ref{fig:bcfl} and Eq.~\eqref{eq:stocfl}. This knowledge-sharing scheme allows the data information to be transferred across cluster models to enhance performance as illustrated in Figure~\ref{fig:stocfl-highlevel}.

In summary, our contributions are proposing a practical CFL framework termed StoCFL. It is flexible with the typically federated learning protocols, making CFL algorithms powerful against various Non-IID issues. We conduct a comprehensive evaluation of the StoCFL framework across four Non-IID settings and real-world datasets (FEMNIST, FedCelebA), encompassing both cross-device scenarios (up to 4,800 clients) and cross-silo setups (20 clients). Experimental results demonstrate that SFCC achieves robust clustering performance under diverse Non-IID conditions, while cluster models trained with Bi-CFL consistently outperform conventional CFL methods in most experiments. Experimental findings confirm that StoCFL surpasses baseline CFL frameworks, offering superior generalization performance and greater system flexibility.

\section{Background \& Related Work}
\textbf{Non-IID Data Issues}. 
FedAvg~\citep{mcmahan2017communication} effectively addresses decentralized data optimization problems when the decentralized data are independently and identically distributed (IID). However, in real-world federated learning, the data distributions are usually Non-IID. In this case, FedAvg would take more rounds to iterate the FL model under the Non-IID data setting. The theoretical analysis~\citep{zhao2018federated} on Non-IID data reveals that the local gradients from FL clients are in significant variance and divergence in direction.
It consequently damages the convergence performance of corresponding optimization algorithms~\citep{zhang2024faster}. Following this study~\citep{hsieh2020non}, we review four Non-IID scenarios that widely exist in real-world applications. They are summarized as follows:
\begin{itemize}
    \item \textit{Feature distribution skew}: The marginal distribution of label $\mathcal{P}(x)$ varies across clients. In a handwriting recognition task, the same words written by different users may differ in stroke width, slant, etc.
    \item \textit{Label distribution skew}: The marginal distribution of label $\mathcal{P}(y)$ may vary across clients. For instance, clients' data samples are tied to geo-regions - kangaroos are only in Australia or zoos; for an online shopping website, customers only buy certain items with specific preferences.
    \item \textit{Feature concept skew}: The condition distributions $\mathcal{P}(x|y)$ varies across clients, and $\mathcal{P}(y)$ is the same. In other words, the same label $y$ can indicate different features $x$ in different FL clients. For instance, the images of items could vary widely at different times and with different illumination. Hence, the data samples with the same label could look very different.
    \item \textit{Label concept skew}: The condition distributions $\mathcal{P}(y|x)$ varies across clients and $\mathcal{P}(x)$ is the same. Similar feature representations from different clients could have different labels because of personal preferences, and the same sentences may reflect different sentiments in language text. 
\end{itemize}
Furthermore, data samples in clients may be a hybrid distribution of the above scenarios in real-world situations, posing a new and formidable challenge to FL techniques. 


\textbf{Clustered Federated Learning}. 
To tackle these Non-IID data issues, conventional federated learning algorithms ~\citep{li2020federated, DBLP:conf/iclr/AcarZNMWS21,  oh2022fedbabu, ega2205} try to learn a global model that generalizes to all distributions. Another line of research~\citep{li2021ditto, marfoq2021federated, t2020personalized, fan2023personalized} focuses on learning personalized models for each client. In contrast, clustered federated learning (CFL) is a new framework for dealing with Non-IID issues in federated learning, representing a middle ground between conventional federated learning and personalized federated learning. 
The first CFL framework is proposed in~\citep{sattler2020clustered, sattler2020byzantine}. Other studies follow the CFL-based framework~\citep{ghosh2020efficient, briggs2020federated, duan2021flexible}. 
The core assumption of CFL is described in Assumption~\ref{asp:cfl}. Clients with comparable data-generating distributions should be clustered, according to this assumption. This clustering ensures that conventional machine learning assumptions (e.g., IID datasets) are satisfied by the decentralized local datasets in the same client cluster. 
Then, it intends to train a shared personalized model for each cluster of clients to mitigate the impacts of Non-IID data issues. And, the model performance can be improved by an intra-cluster federated learning procedure.

CFL frameworks typically include a \textit{federated client clustering} procedure and a \textit{federated cluster model training} procedure. Federated client clustering is a technique for identifying and grouping clients with similar data distributions. For example, Works~\citep{stallmann2022towards, duan2021flexible, pedrycz2021federated, wang2022federated, xie2020multi} clusters clients using K-means clustering based on client parameters. CFL~\citep{sattler2020clustered, sattler2020byzantine} separates clients into two partitions, where clients' gradient directions in the same partition are congruent. FL+HC~\citep{briggs2020federated} uses local updates to produce hierarchical clustering. IFCA~\citep{ghosh2020efficient}/HypCluster~\citep{mansour2020three} implicitly clusters clients by broadcasting different models to them and allowing them to choose which cluster to join based on local empirical loss (hypothesis-based clustering). Their experimental results have demonstrated that client clustering improves the convergence performance of FL models via reducing gradient variance. For the cluster model training procedure, conventional CFL~\citep{briggs2020federated, sattler2020clustered, sattler2020byzantine} train cluster models for each cluster using FedAvg. IFCA~\citep{ghosh2020efficient} conducts parameters-sharing in feature extractor layers and trains personalized output layers via FedAvg. 


\section{Methodology of StoCFL}

This section introduces the StoCFL framework which consists of SFCC and Bi-CFL algorithms. We formally describe the vanilla CFL framework objective and analyze its main drawbacks in Section \ref{sec:pre}. Then, we propose a stochastic client clustering algorithm in Section \ref{sec:clientclustering}. Subsequently, we present the bi-level CFL algorithm aimed at training better cluster models in Section \ref{sec:clusteredlearning}.


\subsection{Problem Formulation}\label{sec:pre}

The conventional CFL framework~\citep{sattler2020clustered} clusters clients with similar data distribution to alleviate data heterogeneity impacts on model optimization. Concretely, CFL minimizes the following objectives for each cluster $k\in[K]$ respectively:
\begin{eqnarray} \label{eq:cfl}
     & \min_{\boldsymbol{\theta}_k} & f_k(\boldsymbol{\theta}_k) =  \mathbb{E}_{\mathcal{D}^{(k)} \sim \mathcal{P}_k}[\mathcal{L}(\boldsymbol{\theta}_k; \mathcal{D}^{(k)})]  \\
    & \mathrm{s. t.} & \; \mathcal{C} = \{c_1, \dots, c_K\}, \; \mathcal{D}^{(k)} = \cup_{i\in c_k} \mathcal{D}_i,\nonumber
\end{eqnarray}
where $\mathcal{L}$ denotes the loss function, $\mathcal{C}$ denotes client clustering results, $c_i \in \mathcal{C}$ denote the set of client indices, and $\mathcal{D}^{(k)}$ denotes the data samples of all clients in cluster $k$ which is sampled from a latent distribution $\mathcal{P}_k$. Notably, $f_k(\boldsymbol{\theta}_k) = \frac{1}{|c_k|}\sum_{i\in c_k} f_i(\boldsymbol{\theta}_k) = \frac{1}{|c_k|}\sum_{i\in c_k} \mathbb{E}_{\mathcal{D}_i \sim \mathcal{P}_k}[\mathcal{L}(\boldsymbol{\theta}_k; \mathcal{D}_i)]$ denotes cluster empirical loss in the $k$-th cluster. Besides, we use $[R]$ to denote the set of integers $\{1,\dots,R\}$, $\|\cdot\|$ to denotes the $\ell_2$ norm of vectors, and $|\cdot|$ to denote the size of a set. For FL notation, we let $N$ be the number of all federated clients. We use $\mathcal{P}_k, k\in[K]$ to denote the latent data source distribution, and $K$ is the number of data source distributions. We consider each federated client $i \in [N]$ to have a local data set $\mathcal{D}_i$. 

We note that the vanilla CFL framework performance is subjected to the client clustering results and corresponding cluster models' training. Here, we emphasize two potential problems hindering the vanilla CFL framework, that we intend to solve:

\textbf{Vanilla client clustering method is not system flexible.} 
In cross-device federated learning, the server may not communicate with all clients simultaneously, because some clients may be running local works or in bad communication conditions. Hence, the server typically randomly samples a subset of clients to run a federated learning process for each communication round. Therefore, we need an iterative federated client clustering procedure to match the vanilla federated learning protocol for the best system flexibility.

\textbf{Vanilla cluster models' training suffers from low data efficiency.}
In the vanilla CFL framework of Eq.~\eqref{eq:cfl}, cluster models are trained using data samples within the same clusters, where the data information of different clusters is not fully communicated. As a result, data information is not transferred among clusters, leading to low data efficiency for model training. Hence, there is a need for a data-effective CFL framework to enhance FL performance.

\begin{figure}
    \centering
    \subfigure[Label distribution skew]{\includegraphics[width=0.23\linewidth]{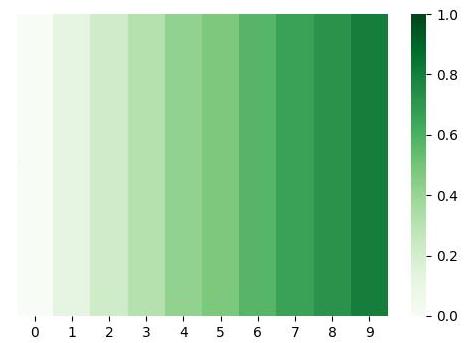}}
    \subfigure[Feature distribution skew]{\includegraphics[width=0.23\linewidth]{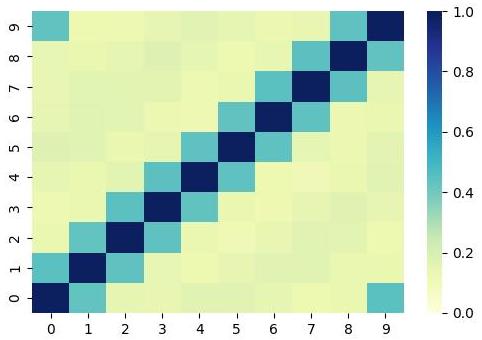}}
    \subfigure[Label concept skew]{\includegraphics[width=0.23\linewidth]{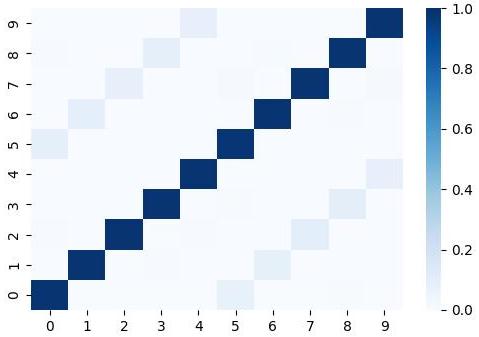}}
    \subfigure[Feature concept skew]{\includegraphics[width=0.23\linewidth]{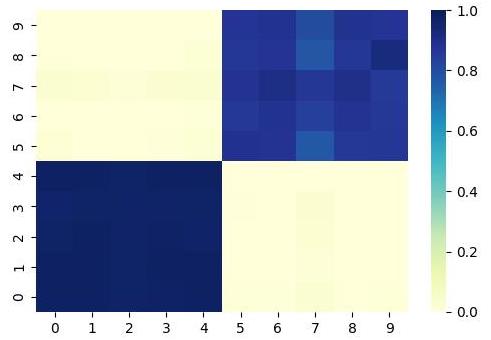}}
    \caption{Visualization of the cosine similarity matrix. Let $s$ denote the value of x-axis or y-axis in the figures. For the Figure (a), $s$ indicates the number of the same label between these two clients. For the Figure (b), we rotated the data with $36\times s$ degrees. For the Figure (c), we modified the label $y = (y+s)$ mod $10$. For the Figure (d), we use gradients calculated based on data samples from MNIST ($s\in[0,5)$) and Fashion-MNIST ($s\in[5,10)$). Then, we random initialize a linear model as anchor $\boldsymbol{\psi}$ for the digit recognition task. We calculate the representation $\Psi(\cdot)$ based on these partitioned datasets and observe the cosine similarity.}
    \label{fig:observe}
\end{figure}

\subsection{Stochastic Federated Client Clustering}\label{sec:clientclustering}

This section introduces the \textit{stochastic federated client clustering} (SFCC) algorithm. SFCC measures the data distribution of clients via a gradient-based function. Then, SFCC clusters clients based on the given objective in Eq.~\eqref{eq:gred}.

\textbf{Local data distribution representation.} 
First, we denote a distribution extractor function $\Psi(\mathcal{D}) = \texttt{Normalize(}\nabla_{\psi} \mathcal{L}(\boldsymbol{\psi}; \mathcal{D})$), which relates the input dataset $\mathcal{D}$, anchor model $\boldsymbol{\psi}$ and loss function $\mathcal{L}$. Theoretically, this function signifies the direction from the anchor model to the local minimum point corresponding to the input dataset $\mathcal{D}$. Intuitively, clients with similar local data distributions exhibit close local minima in the optimization landscape. 
Hence, we anticipate that datasets with similar data distributions yield similar $\Psi(\cdot)$ values. Concretely, we employ the cosine similarity to evaluate the distribution similarity of these two decentralized datasets $\mathcal{D}_i$, $\mathcal{D}_j$:
\begin{equation}
\nonumber
    \cos(\Psi(\mathcal{D}_i), \Psi(\mathcal{D}_j)) = \frac{\Psi(\mathcal{D}_i) \cdot \Psi(\mathcal{D}_j)}{\|\Psi(\mathcal{D}_i)\| \|\Psi(\mathcal{D}_j)\|}.
\end{equation}
To justify this metric, we conduct observational experiments on the cosine similarity, as depicted in Figure~\ref{fig:observe}. The results indicate a significant difference in cosine similarity values and their relation between clients' datasets. In other words, the experiments demonstrate that clients with similar local data distributions exhibit higher cosine similarity at both label and feature levels. Therefore, the output of $\Psi(\cdot)$ is interpreted as a representation of the data distribution corresponding to the input dataset. Moreover, clients with similar local data distribution representation preserve consistent gradients on the model~\citep{li2023revisiting, sattler2020clustered}. A group of federated clients preserves similar $\Psi(\cdot)$ values indicating low gradient variance across clients(less Non-IID impacts), which benefits intra-cluster model convergence.

\textbf{SFCC objective.} 
Based on the observations, we propose a stochastic federated client clustering algorithm that clusters federated clients by minimizing the following objective: 
\begin{equation}\label{eq:gred}
\begin{aligned}
    \min_{\mathcal{C}}  & \; \sum_{i=1}^{\tilde{K}} \sum_{j=i+1}^{\tilde{K}}  \cos\big(\Psi(\tilde{\mathcal{D}}^{(i)}), \Psi({\tilde{\mathcal{D}}^{(j)}})\big), \\ 
\end{aligned}
\end{equation}
where $\Psi({\tilde{\mathcal{D}}^{(j)}}) \triangleq \frac{1}{|c_j|}\sum_{i\in c_j}\Psi(\mathcal{D}_i)$ is the average point for decentralized local datasets of clients in the current $j$-th cluster in $\mathcal{C}$, $\tilde{K}$ denotes the number of clusters and $\tilde{K} = |\mathcal{C}|$. Importantly, we use a threshold $\tau$ to indicate the minimum cosine similarity that two datasets should be considered similar. In other words, minimizing this objective aims to identify the optimal partition $\mathcal{C}$ where the representations of current clusters are maximally distant from each other.

\textbf{SFCC outline.}  
At the initialization stage, we regard each client as a single cluster. In other words, the server maintains a partition set $\mathcal{C} = \{c_1, c_2, \dots, c_N\}$,  where $N$ is the number of clients. For each $c_i \in \mathcal{C}$, we have $c_i = \{i\}$ and $i$ denotes the client id. Meanwhile, we have $\Psi(\tilde{\mathcal{D}}^{(k)}) = \Psi(\mathcal{D}_i)$ for all $k,i \in [N]$ and $\tilde{K}=N=|\mathcal{C}|$ at the beginning. 
For the federated client clustering process, we greedily decrease the value of Equation \eqref{eq:gred} by merging the similar clusters sampled at each round. In practice, the Equation \eqref{eq:gred} can be represented by a pair-wise cosine similarity matrix $M$, where $M_{i,j}$ indicates the cosine similarity between the $i$-th and the $j$-th clusters in $\mathcal{C}$. 
We summarize the stochastic federated client clustering procedure in Lines 4-13, Algorithm~\ref{ifc}. For each federated round, the server would request the local data distribution representation from the sampled clients (Line 6). Then, the server updates the distribution representation of clusters (Line 9). Finally, the server merges any two clusters that satisfy the requirements (Lines 11-13). For cluster merging operation, the current number of clusters $\tilde{K}$ is reduced by 1. If the federated server samples all clients at the first round, StoCFL recovers to client-wise agglomerative clustering, with the metric provided by distribution extractor $\Psi(\cdot)$.

\subsection{Bi-level Clustered Federated Learning}\label{sec:clusteredlearning}

This section introduces a bi-level CFL objective to enhance conventional CFL frameworks further. While Eq.~\eqref{eq:cfl} outlines the objective of conventional CFL, where cluster models are optimized independently within each cluster, our proposed bi-level CFL objective recognizes the potential benefits of inter-cluster knowledge sharing. Despite potential differences in the implicit data distribution between clusters, we justify that certain knowledge from one cluster model can improve the performance of another.

Specifically, aligned with the client clustering procedure, the server maintains a global model denoted by $\boldsymbol{\omega}$ and cluster models $\boldsymbol{\theta}_k, k\in[\tilde{K}]$. Our approach addresses a bi-level optimization problem for each cluster $k\in [\tilde{K}]$:
\begin{equation}\label{eq:stocfl}
\begin{aligned}
   \min_{\boldsymbol{\theta}_k}  \; &  h_k(\bs{\theta}_k; \bs{\omega}^*) := f_k(\boldsymbol{\theta}_k) + \frac{\lambda}{2}\|\boldsymbol{\theta}_k - \boldsymbol{\omega}^*\|^2 \\ 
    \mathrm{s. t.}\;  & \boldsymbol{\omega}^* \in \arg \min_{\boldsymbol{\omega}} G\big(f_1(\boldsymbol{\omega}), \dots, f_N(\boldsymbol{\omega})\big), \\
    & \mathcal{C} = \{c_1, \dots, c_K\}, \; \mathcal{D}^{(k)} = \cup_{i\in c_k} \mathcal{D}_i,
\end{aligned}
\end{equation}
where $N$ represents the total number of clients, $f_i(\cdot) = \mathbb{E}[\mathcal{L}(\cdot; \mathcal{D}_i)], i\in[N]$ denotes the empirical loss on the $i$-th client, and $G(\cdot)$ denotes the global objective function for training the global model $\boldsymbol{\omega}$. In this paper, we adopt $G(\cdot)$ as FedAvg. The global model $\bs{\omega}^*$ preserves general data information as it is trained on whole datasets across clients. The regularization implicitly enhances the data efficiency for the training of cluster model~\citep{li2021ditto, wengprobabilistic}.

We propose to solve Eq.~\eqref{eq:stocfl} through alternating training between global and cluster models. Aligned with the stochastic federated client clustering procedure, we initialize $\boldsymbol{\omega}_0 = \boldsymbol{\theta}_1 = \dots = \boldsymbol{\theta}_N$ and $\tilde{K} = N$. Additionally, if any two clusters are merged during the client clustering process, the server merges the corresponding cluster models. Therefore, the real-time number of clusters $\tilde{K}$ always matches the number of cluster models. The pseudocode for bi-level CFL is outlined in Lines 14-23 of Algorithm \ref{ifc}. Throughout the training process, the server broadcasts the global model $\boldsymbol{\omega}$ and the corresponding cluster model $\boldsymbol{\theta}_k$ to sampled clients. Subsequently, the sampled clients execute several steps of SGD to locally optimize the cluster model (Line 21) and global model (Line 22) before uploading the updated models to the server. The server updates the global model by aggregating the models from all sampled clients (Line 17) and then updates the cluster models (Lines 18-19).

\begin{center}
\scalebox{0.9}{
\begin{minipage}{0.9\linewidth}
\begin{algorithm}[H]
\SetAlgoVlined
\IncMargin{1em}
\KwIn{Client set $S$, where $|S| = N$, initialized cluster partition $\mathcal{C} = \{c_1, \dots, c_N\}$, initialized model $\boldsymbol{\omega}_0, \boldsymbol{\theta}=[\boldsymbol{\theta}_1, \dots, \boldsymbol{\theta}_N]$, anchor $\boldsymbol{\psi}$, threshold $\tau$ and learning rate $\eta$.}
\KwOut{Cluster result $\mathcal{C} = \{c_1, \dots, c_{\tilde{K}}\}$, cluster models $\boldsymbol{\theta}_1, \dots, \boldsymbol{\theta}_{\tilde{K}}$, and global model $\boldsymbol{\omega}$}
\SetKwFunction{FMain}{ServerProcedure}
\SetKwFunction{CMain}{ClientProcedure}
\SetKwProg{FPc}{}{:}{}
\SetKwProg{Pc}{}{($\boldsymbol{\omega}, \boldsymbol{\theta}_k$):}{}
\FPc{\FMain}{
    $P \leftarrow \emptyset$ 
    
    \For{round $t$ $\in$ $[T]$}{
        Random sample a subset of client $S^t \subseteq S, |S^t| = m$. 
        
        \tcp{federated client clustering}
        \For{client id $i \in (S^t \cap \complement_S P)$ in parallel}{
            $\Psi(\mathcal{D}_i) \leftarrow \textrm{GetDatasetRepresentation}(\boldsymbol{\psi})$ 
        }
        
        $P \leftarrow P \cup S^t$
        
        \For{$c_k \in \mathcal{C}$}{
            $\Psi(\tilde{\mathcal{D}}^{(k)}) = \sum_{i\in c_k} \Psi(\mathcal{D}_i)$
        }
        
        Obtain cosine similarity matrix $M$ following Eq.~\eqref{eq:gred}.
        
        
        \For{$M_{i,j}$ in $M$}{
            \If{$M_{i,j} > \tau$}{
                $c_i \leftarrow c_i \cup c_j, \mathcal{C} \leftarrow \mathcal{C}/c_j$
            }
        }

        \tcp{clustered federated learning}
        \For{client id $i \in S^t$ in parallel}{
            Let $k$ denote the index of the cluster and client id $i \in c_k$.
            
            $\boldsymbol{\omega}^i_t, \boldsymbol{\theta}_k^i \leftarrow \textrm{ClientProcedure}(\boldsymbol{\omega}_t, \boldsymbol{\theta}_k)$
        }
        $\boldsymbol{\omega}_{t+1} \leftarrow \textrm{FedAvg}([\boldsymbol{\omega}^i_t]), i \in S^t$
        
        \For{$c_k \in C^t$ and indexed by $k$}{
            $\boldsymbol{\theta}_k \leftarrow \textrm{FedAvg}([\boldsymbol{\theta}_k^i]), i \in c_k$
        }
    }
}

\Pc{\CMain}{

    $\boldsymbol{\theta}_k^i \leftarrow \boldsymbol{\theta}_k -  \eta\big(\nabla f_i(\boldsymbol{\theta}_k) + \lambda(\boldsymbol{\theta}_k - \boldsymbol{\omega})\big)$
    
    $\boldsymbol{\omega}^i  \leftarrow \boldsymbol{\omega} - \eta \nabla f_i(\boldsymbol{\omega})$

    return $\boldsymbol{\omega}^i, \boldsymbol{\theta}_k^i$
}
\caption{Stochastically Clustered Federated Learning}
\label{ifc}
\end{algorithm}

    \end{minipage}%
    }
  \end{center}

\

\section{Discussion on Implications of StoCFL}\label{sec:discuss}

In this section, we discuss the main implications of StoCFL and provide further insights into its methodology.

\textbf{Flexible stochastic federated client clustering (SFCC).} 
The SFCC algorithm is compatible with standard federated client sampling protocols. 
Previous K-means-based~\citep{stallmann2022towards, duan2021flexible, pedrycz2021federated, wang2022federated, xie2020multi}, or hierarchical-clustering-based client clustering~\citep{sattler2020byzantine, sattler2020clustered} typically requires once full client communication before federated learning process to obtain client clustering results. 
In contrast, SFCC does not require all clients to participate in any communication round. Moreover, one additional communication is necessary to download the anchor model and upload the local data distribution representation vector. In comparison, hypothesis-based client clustering~\citep{ghosh2020efficient, mansour2020three} typically broadcasts multiple cluster models to individual clients, resulting in additional communication overhead, especially during frequent communication instances.

\textbf{Comparison between the vanilla CFL framework in Eq.~\eqref{eq:cfl} and StoCFL framework in Eq.~\eqref{eq:stocfl}.} 
The proposed StoCFL framework \textit{enhances cluster models with shared knowledge from global model $\boldsymbol{\omega}$}. In our bi-level optimization objective, the global model $\boldsymbol{\omega}$ is trained on the data samples of all clients. Hence, the global model preserves intensive knowledge (e.g., averaged data distribution and feature information) among clients~\citep{hoang2024few, wengprobabilistic}.
As shown in Figure~\ref{fig:stocfl-highlevel}, the global regularization term in cluster model optimization allows knowledge from isolated clusters to be transferred to others through the global model. Furthermore, the cluster model training operates independently of the global model training process, rendering StoCFL flexible in selecting the global objective $G(\cdot)$. As a result, the cluster model could inherit convergence benefits and model performance from the global model, such as robustness, fairness~\citep{li2021ditto}, and generalization performance. 
In edge cases, large $\lambda$ restricts cluster models closer to the global model, which guarantees the cluster models will not be worse than the global model.
When $\lambda$ is set to 0, the StoCFL returns to the conventional CFL framework described in Eq.~\eqref{eq:cfl}. 
In practice, tuning the regularization weight $\lambda$ adjusts the benefits that cluster models gain from the global model, as shown in Table~\ref{tab:effectl}, Section~\ref{sec:hyper}. 

\textbf{Optimization benefits from clustering results $\mathcal{C}$.} 
SFCC builds client clusters relying on the cosine similarity of local gradients, which is a better client clustering metric than model-parameters~\citep{ghosh2020efficient, mansour2020three}.
Note that cosine similarity between clients' gradients indicates the similarity of local datasets~\citep{zeng2023power, li2023revisiting}, reflecting the data heterogeneity across clients. 
SFCC clusters clients with high cosine similarity lead to a smaller intra-cluster gradient variance. The reduced 
intra-cluster gradient variance enhances the performance of cluster models.
Notably, the client clustering results, represented by $\tilde{K}$ and $\mathcal{C}$, depend on the merging threshold $\tau$. Specifically, no clusters will merge if $\tau=1$, resulting in a final number of clusters $\tilde{K} = N$, thereby simplifying the StoCFL framework to the personalized FL algorithm Ditto~\citep{li2021ditto}. Conversely, if $\tau=-1$, all clients will be clustered together, leading the optimization objective to degenerate into the global FL algorithm FedProx~\citep{li2020federated}. 
In practice, adjusting the clustering threshold $\tau$ can effectively balance between global FL and personalized FL.

\textbf{Cluster inference for newly joined clients}. 
StoCFL demonstrates flexibility with newly joined clients. Specifically, it adapts to inferring a newly joined client during or after the training process. For instance, consider a newly joined client with a local dataset $\hat{\mathcal{D}}$ reporting the local representation $\Psi(\hat{\mathcal{D}})$ to the server. The server can then determine the target cluster through the following steps:
\begin{enumerate}
    \item Calculate the candidate cluster with the closest distance 
    $$\hat{\mathcal{D}} = \arg \max \text{cos}(\Psi(\hat{\mathcal{D}}), \Psi(\tilde{\mathcal{D}}^{(j)})), \; j\in [\tilde{K}],$$ 
    and record the candidate cluster id $\hat{c}$.

    \item If $\hat{\mathcal{D}} \geq \tau$, then assign the client to the cluster $\hat{c}$. Otherwise, the server assigns the client to a new cluster marked by $\tilde{K}+1$, $\tilde{K} = \tilde{K}+1$, and let $\theta_{\tilde{K}} = \theta_{\hat{c}}$.
\end{enumerate}
Additionally, if the client is unable to join any existing cluster, the server assigns a cluster model with the closest cluster distribution. This facilitates learning from a good initialization during the training process. Furthermore, after completing the FL training process, we can use the closest model to predict data samples from new clients. We evaluate this property in Section~\ref{sec:inference}.

\section{Experimental Evaluation}
In this section, we evaluate the performance of StoCFL using Non-IID settings following previous works~\citep{sattler2020clustered, ghosh2020efficient} and the real-world setting FEMNIST and FedCelebA~\citep{caldas2018leaf}. Our experiment investigation includes federated client clustering and CFL evaluation. Furthermore, we discuss the effect of StoCFL hyper-parameters. Then, we demonstrate the application inference ability of StoCFL and study its generalization ability to unseen clients. The experiment is implemented using the open-source framework~\citep{zeng2023fedlab, liang2023fednoisy} and the source code is publicly available~\footnote{https://github.com/dunzeng/StoCFL}.

\subsection{Stochastic Client Clustering with Non-IID Data}\label{sec:stoclu}

We evaluate the stochastic federated client clustering algorithm on the MNIST~\citep{lecun1998gradient} and Fashion-MNIST~\citep{li2021federated}. We recall that the MNIST/Fashion-MNIST dataset has 60,000 training samples and 10,000 test samples with ten classes. To simulate an FL environment where the data distributions among clients are different, we augment and partition these datasets as follows.
\begin{itemize}
    \item \textbf{Pathological MNIST}~\citep{mcmahan2017communication}: we sort the data samples by labels and split them into \{\{0,1,2\},\{3,4\},\{5,6\},\{7,8,9\}\} (4 clusters). Then we randomly partition the dataset into 100 clients for each cluster, resulting in 400 clients with $K=4$. We refer to it as \textit{Label distribution skew}.
    \item \textbf{Rotated MNIST}~\citep{lopez2017gradient}: we augment MNIST by rotating images with 0, 90, 180, and 270 degrees, resulting in $K=4$ clusters. We randomly partition the Rotated MNIST of each degree into 100 clients. This scenario is \textit{Feature distribution skew}.
    \item \textbf{Shifted MNIST}~\citep{sattler2020clustered}: we modify the label of each sample by adding shift level $s$(i.e., $\bar{y}_s = (y+s)\%10, s\in{0,3,6,9}$) and partition dataset of each shift level into 100 clients. Hence, this case shall be \textit{Label concept skew}.
    \item \textbf{Hybrid MNIST}: we partition MNIST and Fashion-MNIST into 100 clients, resulting in 200 clients and 2 clusters. The label across clients is the same ($y \in [10]$), but the feature domain is different. Therefore, this case is \textit{Feature concept skew}.
\end{itemize}

\begin{figure}
    \centering
    \includegraphics[width=\linewidth]{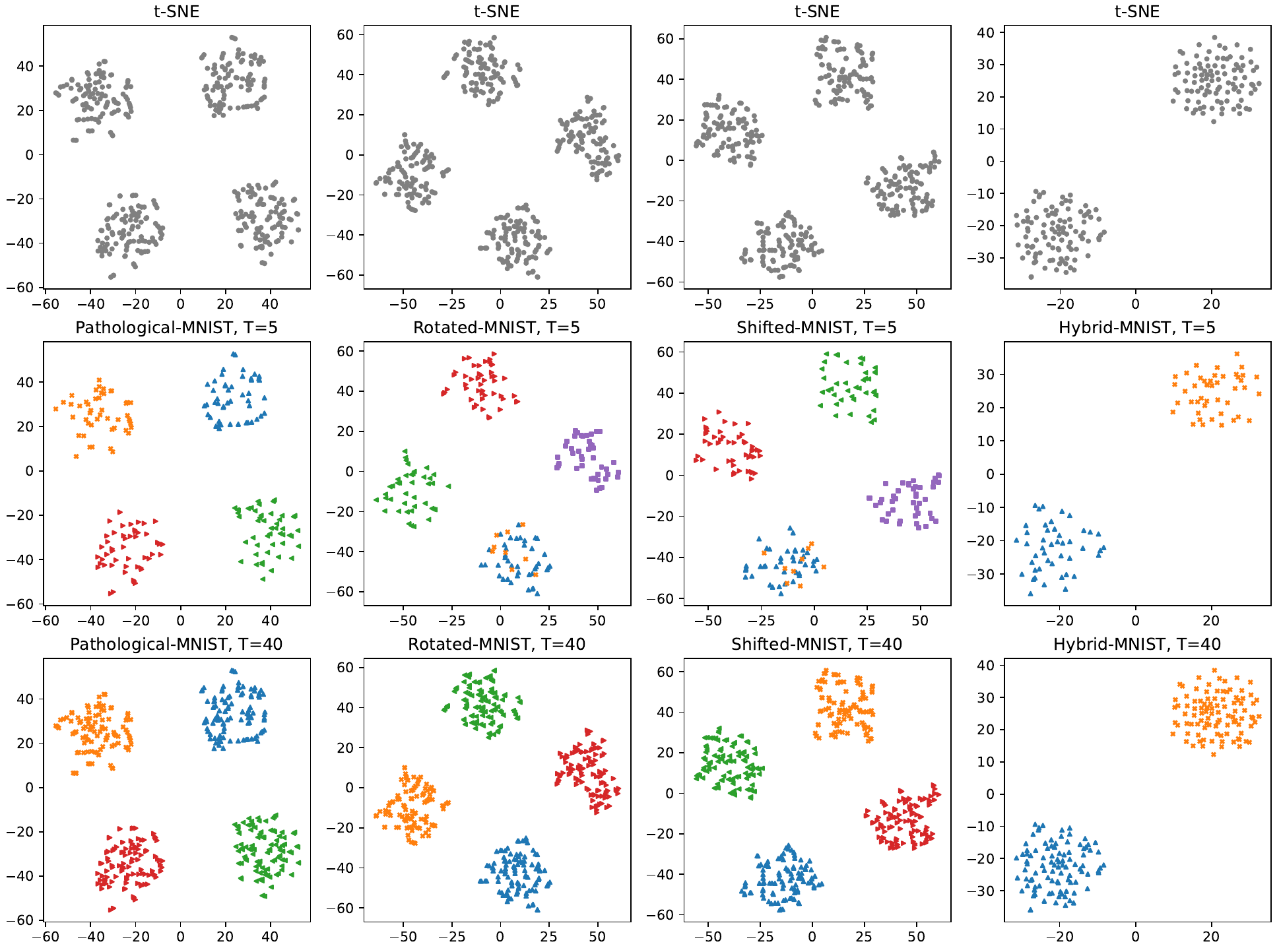}
    \caption{Illustration of stochastic client clustering. The first row is the t-SNE results on all $[\Psi(\mathcal{D}_i)], i\in[N]$, where each point denotes the data distribution of the dataset from each client. The other row is the clusters at communication round $T = \{5,10,40\}$, which are colored by SFCC.}\label{visual1}
\end{figure}

\begin{figure}
    \centering
    \includegraphics[width=\linewidth]{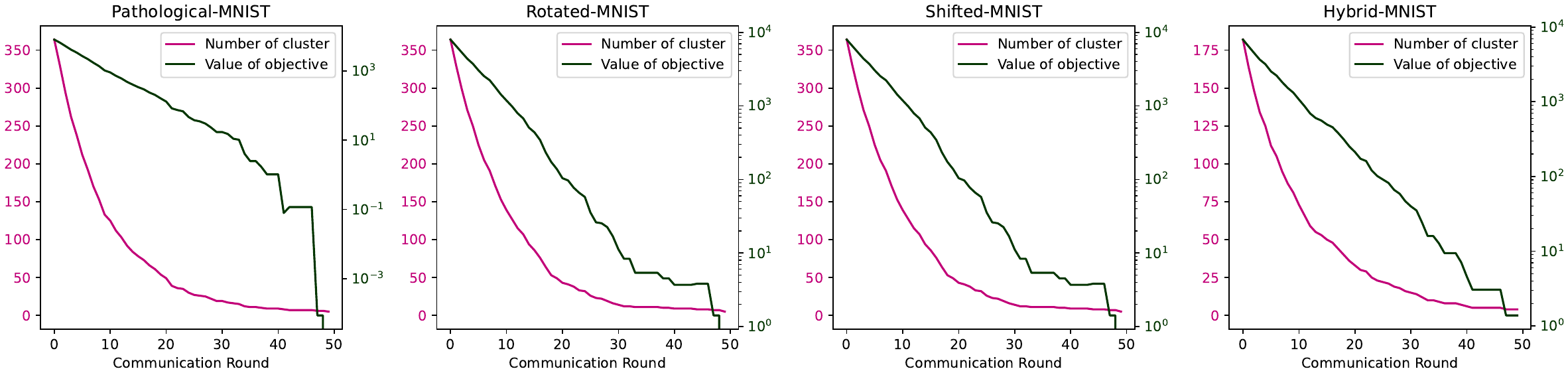}
    \caption{The overall number of clusters and the value of the clustering objective in Equation~\ref{eq:gred} at each communication round.}\label{visual2}
\end{figure}

In experiments, we randomly initialize an anchor model $\boldsymbol{\psi}$ and choose the cross-entropy loss as $\mathcal{L}$ for the classification task. Based on that, we obtain all representation vectors using $\Psi(\cdot)$ and display them using t-SNE~\citep{van2008visualizing} visualizing. We conducted 50 rounds of the proposed stochastic federated client clustering procedure with $\tau=0.5$, and a sample rate of 10\%. 

The visualization results are shown in Figure~\ref{visual1} and~\ref{visual2}. In Figure~\ref{visual1}, the graph with grey points depicts the visualization results of distribution representation vectors. Furthermore, clear clusters are shown from the representation vectors captured by t-SNE, which further proves that $\Psi(\cdot)$ could extract and represent the local data distributions. Besides, an additional clustering status curve is depicted in Figure~\ref{visual2}. The curve shows that SFCC minimizes the objective~\ref{eq:gred} and clusters clients into the correct number of clusters without knowing any prior information about clusters.

\subsection{Clustered Federated Learning Performance}\label{sec:cfl_exp}

We evaluate the model performance of StoCFL in this part. 

\textbf{Baselines}. 
We consider FedAvg~\citep{mcmahan2017communication}, FedProx~\citep{li2020federated}, and FedInit~\citep{sun2024understanding} as the baseline of the global model. In particular, FedInit is an advanced variant of FedAvg, which designs a personalized initialization for local models at every communication round. This procedure has been proven to be fast convergence and better generalization.
Moreover, we also consider the personalized federated learning algorithm Ditto~\citep{li2021ditto} as the baseline of cluster models. For CFL baselines, we compare the IFCA~\citep{ghosh2020efficient}, CFL~\citep{sattler2020clustered}, and FedSEM~\citep{xie2020multi} under the same Non-IID settings reported in their paper. Here, we describe the details of baseline CFL methods and highlight the methodological differences. 

IFCA~\citep{ghosh2020efficient} takes an input of the assumption number of cluster $\tilde{M}$. Then, the server initializes $\tilde{M}$ different models for each cluster and broadcasts them to clients for each round. The clients will optimize the specific model, which achieves the lowest forward loss on the local dataset. Clients upload the updated model to the server. The server would aggregate the updated $\tilde{M}$ models in the same cluster following FedAvg. For ease of writing, we use IFCA($\tilde{M}$) to denote the cluster number inputs.

CFL~\citep{sattler2020clustered} monitors the model updates of all clients for each round. In the beginning, all clients are in the same cluster. Then, the server will bi-partition clients into two clusters when certain conditions are satisfied. The CFL server would recursively run the above procedures in the same cluster until client clustering is finished.

{FedSEM~\citep{xie2020multi} leverage the Euclidean distance between local models and global models. Then, each local model is assigned to its closest global model. Analogous to IFCA, FedSEM also initializes $\tilde{M}$ different global models as the center model of each cluster. For each communication round, the updated local models will be adopted to update their corresponding global center model following the FedAvg protocol. Also, we use FedSEM($\tilde{M}$) to denote the cluster number inputs.}

\textbf{Experiment Setup}. We compare the baseline algorithms using identical experimental conditions described in their study. The MNIST task model is a linear classification model with a hidden layer of 2048 units, and the CIFAR10 task model is a convolution neural network model with two convolutional layers followed by two fully connected layers. We execute five runs for each experiment, each with a different random seed. We also provide the average test accuracy and standard deviation. Without loss of generality, we initialize model $\boldsymbol{\psi} = \boldsymbol{\omega}_0$ for the distribution extractor function $\Psi(\cdot)$ in experiments, and the loss function $\mathcal{L}$ is the cross-entropy loss for the classification task. As experiments are conducted with multiple random seeds, we argue that the distribution extractor $\Psi(\cdot)$ is robust to the model initialization. 

\begin{table*}[t]
\caption{Test accuacies(\%)$\pm$ std.}
\label{tab:ifca}
\resizebox{\linewidth}{!}{
\begin{tabular}{lcccccccc}
\toprule[\heavyrulewidth]
& \multicolumn{6}{c}{\textbf{Rotated MNIST}, K=4} & \multicolumn{2}{c}{\textbf{Rotated CIFAR}, K=2} \\
\cmidrule(r){2-7}
\cmidrule(r){8-9}
$N$, $|\mathcal{D}_i|$         & \multicolumn{2}{c}{4800, 50} & \multicolumn{2}{c}{2400, 100} & \multicolumn{2}{c}{1200, 200} & \multicolumn{2}{c}{200, 500}      \\
\cmidrule(r){2-3} 
\cmidrule(r){4-5} 
\cmidrule(r){6-7} 
\cmidrule(r){8-9}
Sample Rate             & 10\%            & 100\%           & 10\%            & 100\%            & 10\%            & 100\%            & 10\%             & 100\%             \\ \midrule 
FedAvg & 95.72$\pm$0.13 & 96.10$\pm$1.26              & 95.32$\pm$0.16 & 96.00$\pm$1.31              & 94.96$\pm$0.11 &  95.72$\pm$1.60             & 48.26$\pm$0.66    &  47.98$\pm$1.73              \\
{FedInit}     & {96.29$\pm$0.52} & {97.04$\pm$0.38}    & {96.47$\pm$1.16} &    {96.77$\pm$0.47}          &    {96.03$\pm$1.53}      & {96.65$\pm$0.72}             &     {48.09$\pm$0.18}            & {48.79$\pm$0.42}           \\
FedProx &  76.45$\pm$0.07     & 76.47$\pm$0.08    & 77.40$\pm$0.09   & 77.48$\pm$0.08    & 77.82$\pm$0.07   &  77.90$\pm$0.09          & 47.17$\pm$0.35   &  47.15$\pm$0.36              \\
Ditto &  65.05$\pm$0.05     &  73.60$\pm$0.05      &  73.93$\pm$0.11  & 79.71$\pm$0.05   & 79.38$\pm$0.08   &  83.96$\pm$0.03       &  46.91$\pm$0.38   & 46.61$\pm$0.39              \\

IFCA(2)     & 84.42$\pm$1.05 & 84.98$\pm$0.28    & 85.25$\pm$0.94 & 84.64$\pm$2.54             & 86.12$\pm$0.78 &  84.37$\pm$4.40      & 50.62$\pm$1.86  &             49.83$\pm$2.79  \\
IFCA(4)     & 91.74$\pm$0.02 & 90.43$\pm$2.71    & 91.39$\pm$1.24 & 90.75$\pm$2.48            & 91.00$\pm$1.44 &  88.69$\pm$2.83 & 50.80$\pm$1.69  &  51.40$\pm$2.00  \\
IFCA(6)     & 91.74$\pm$0.03 & 91.65$\pm$0.12    & 91.94$\pm$0.13 &    91.99$\pm$0.04          &    92.15$\pm$0.07            & 92.17$\pm$0.04             &     51.04$\pm$0.40            & 50.63$\pm$1.85            \\
{FedSEM(2)}     & {88.11$\pm$0.23} & {89.22$\pm$0.79}    & {89.72$\pm$1.01} &    {90.11$\pm$0.71}          &  {89.67$\pm$0.31}              & {91.67$\pm$0.41}             &     {47.79$\pm$0.71}            & {50.30$\pm$0.17}           \\
{FedSEM(4)}     & {89.04$\pm$1.32} & {90.27$\pm$1.29}    & {90.01$\pm$0.83} &    {90.60$\pm$0.29}          &  {90.92$\pm$0.43}            &  {92.42$\pm$0.67}             &     {46.06$\pm$1.78}            & {48.30$\pm$0.43}           \\
{FedSEM(6)}     & {87.83$\pm$0.15} & {88.62$\pm$0.61}    & {88.43$\pm$1.32} &    {88.92$\pm$1.75}         &  {88.76$\pm$0.27}            &  {89.42$\pm$0.61}             &     {45.03$\pm$2.73}            & {47.13$\pm$1.33}           \\
StoCFL         & \textbf{97.00$\pm$0.04} & \textbf{97.36$\pm$0.38}            & \textbf{96.89$\pm$0.02} & \textbf{97.4$\pm$0.35}             & \textbf{96.71$\pm$0.04} &  \textbf{97.40$\pm$0.38}            & \textbf{52.84$\pm$0.74} & \textbf{54.99$\pm$0.66}       \\
\bottomrule[\heavyrulewidth]
\end{tabular}}
\end{table*}

\subsubsection{{Experiments on Feature-rotated MNIST and CIFAR-10 dataset}}

Following the setting of IFCA, we create the Rotated MNIST and Rotated CIFAR10 datasets. Firstly, We rotated both the training and testing data samples by 0, 90, 180, and 270 degrees, resulting in cluster number $K=4$. Then, we randomly partition training datasets into $N = (4800, 2400, 1200)$ clients, and each client has $|D| = (50, 100, 200)$ data samples. For the MNIST tasks, We run 100 federated communication rounds for the Rotated MNIST setting. Each client performs 5 epochs of local SGD with full batch size and the learning rate $\eta=0.1$. Besides, the Rotated CIFAR10 is created similarly with 0 and 180 degrees of rotation. We run 200 federated communication rounds for the Rotated CIFAR setting. Each client performs 5 epochs of local SGD with a local batch size is 50 and the learning rate $\eta=0.1$. For the hyper-parameters of StoCFL, the stochastic client clustering threshold $\tau=0.5$, and the optimization parameter $\lambda=0.05$. In addition, we compare FedAvg, FedProx, and Ditto to demonstrate the improvement of StoCFL. We report the mean accuracy of cluster models on their corresponding testing datasets and the mean accuracy of the global model on all testing datasets. We highlight that the original experiments in IFCA only report a full client sample rate in MNIST and a client sample rate of 10\% in CIFAR10. For a fair comparison, we fulfill the sample rates for both MNIST and CIFAR10 tasks. 

\textbf{Comparison with IFCA {and FedSEM}.} 
The performance results are shown in Table \ref{tab:ifca}. 
According to the results, StoCFL outperforms IFCA and FedSEM in most cases. More importantly, we observe that IFCA fails to cluster clients from different data distributions in MNIST experiments, where a particular model may dominate another model from the beginning. That is, if a model fits two distributions well in the first few rounds, then this model will dominate another model, making it no chance to fit the expected distribution (no clients will update this model). {Analogous to IFCA, we observed the same phenomenon in FedSEM experiments.} Hence, we argue that IFCA and FedSEM depend on model initialization. Meanwhile, previous studies~\citep{wu2022motley,ma2022convergence} on the heterogeneity of FL observe a similar phenomenon in the IFCA algorithm. {Moreover, we observe that IFCA and FedSEM implement promising clustering results in CIFAR10 tasks. And hence, they outperform global baselines such as FedAvg. However, they lack mechanisms that allow the knowledge to transfer across clusters. Therefore, they achieve lower accuracies than StoCFL.}

\textbf{Comparison with FedAvg, FedInit, FedProx and Ditto.} 
We observed that StoCFL performs better than FedProx and Ditto on both Rotated MNIST and CIFAR experiments. Moreover, the FedProx and Ditto performances are worse on the MNIST tasks, where the number of clients is large and the data samples of each client are fewer. Besides, please refer to our discussion in Section~\ref{sec:discuss}, StoCFL additionally leverages the client clustering results on top of Ditto, and achieves better performance. Therefore, the results show the benefits of conducting client clustering on cross-device federated learning. 

\begin{figure*}[t]
\centering
\begin{adjustbox}{valign=t}
\begin{minipage}[t]{0.45\linewidth}
\centering
\includegraphics[scale=0.5]{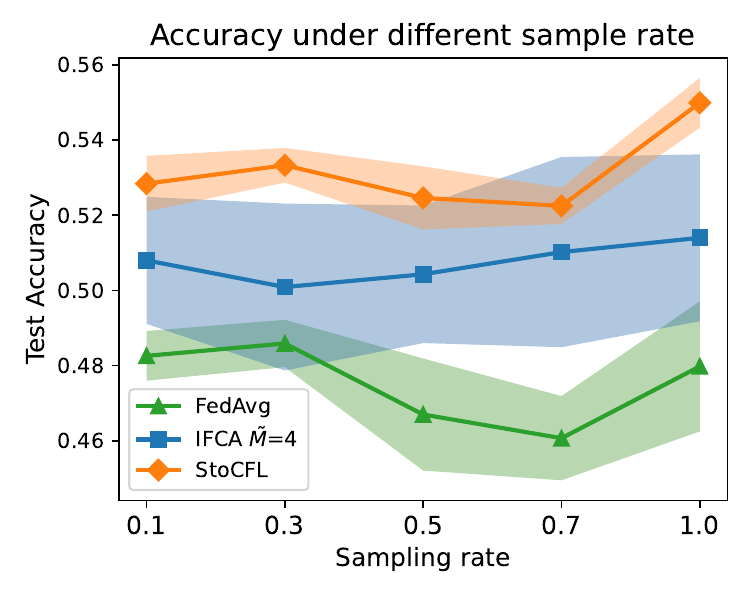}
\caption{Stability comparison of different federated learning methods with various sampling rates.}\label{fig:sample_rate}
\end{minipage}
\end{adjustbox}
\hspace{5pt}
\begin{adjustbox}{valign=t}
\begin{minipage}[t]{0.45\linewidth}
\centering
\includegraphics[width=\textwidth]{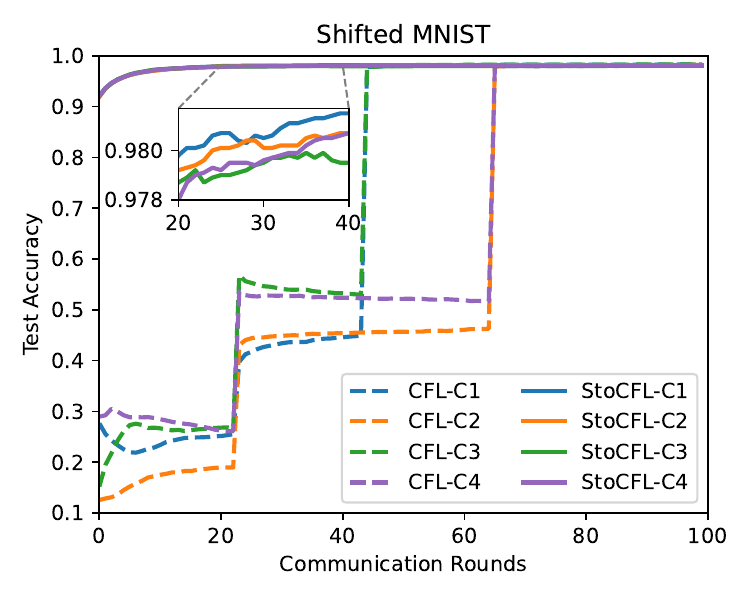}
\caption{Evolution curves of test accuracy for CFL and StoCFL with increasing communication rounds.}
\label{fig:compare_with_cfl}
\end{minipage}
\end{adjustbox}
\end{figure*}

\textbf{Stability.} We note that IFCA requires a prior assumption on the number of clusters $\tilde{M}$. In Table~\ref{tab:ifca}, the results show that IFCA is sensitive to the $\tilde{M}$. Moreover, results in Figure \ref{fig:sample_rate} study the effect of client sampling rate on StoCFL with the Rotated CIFAR setting. The performance curve of StoCFL is stable and better with different settings of the sample rate. Hence, the results reveal that StoCFL is robust and flexible with the proportion of client participation.

\subsubsection{{Experiments on Label-shifted MNIST and CIFAR-10 dataset}}
Following the setting of CFL~\citep{sattler2020clustered}, we conduct experiments on the Shifted MNIST and the Shifted CIFAR-10 dataset \citep{krizhevsky2009learning}. In this case, datasets are partitioned into $N=20$ clients, each belonging to one of $K=4$ clusters. The labels of each client data are modified by randomly shifted labels, i.e., $\tilde{y} = (y+s)\%10, s\in\{0,3,6,9\}$. For a fair comparison, we let all clients participate in StoCFL, and we also conducted {FedInit}, IFCA ($\tilde{M}=4$) and {FedSEM ($\tilde{M}=4$)} in this part. Here, we set the local batch size to 128 and the learning rate $\eta=0.1$.

{The performance is shown in Table~\ref{tb:cfl-results}, where we let the sampling rate be 1. In this case, we observed that conventional FL algorithms (FedAvg, FedInit) performance is highly degenerated as data labels are shifted across clients. While CFL algorithms achieve comparable model performance. This is because label-shift data are easily distinguishable from all aspects including loss (IFCA), model parameters (FedSEM), and local updates (CFL, StoCFL). Meanwhile, as the number of clients is small, the data are more centralized than experiments from Table~\ref{tab:ifca}. Hence, all CFL experiments achieve higher test accuracy than conventional FL algorithms (FedAvg, FedInit).}
Despite that, we note that the accuracy curve of StoCFL is better in most training rounds than CFL as shown in Figure~\ref{fig:compare_with_cfl}. This is because StoCFL finishes client clustering at an early stage, while the CFL requires full client participation for every round to bi-partition the clients at later stages. In contrast, StoCFL supports an arbitrary proportion of client participation, which reveals the flexibility of StoCFL.

\subsubsection{{Evaluation on real-world datasets FEMNIST and CelebA}} 
We further evaluate StoCFL on the real-world dataset Federated Extended MNIST (FEMNIST)~\citep{cohen2017emnist} {and Federated CelebA (FedCelebA)~\citep{liu2015faceattributes}} from LEAF~\citep{caldas2018leaf}. In detail, FEMNIST is a realistic FL dataset where the data samples on a single client are the handwritten digits or letters from a specific writer. Besides, There are 3,597 clients in FEMNIST, and the number of data samples among these clients is different. {In FedCelebA, 200,288 images are extracted for 9,343 clients, and an on-device classifier is trained to recognize whether the person smiles or not.}
Hence, these settings could be considered hybrid Non-IID scenarios. Noting that there are no clear clusters for FEMNIST and FedCeleB clients, the writing styles/face features of different people may be clustered. For the model, we use a neural network with two convolutional layers, a max pooling layer after the convolutional layer, and two fully connected layers. We also added a dropout ratio of 0.5 between the convolutional layer and the fully connected layer. The network parameters initialization follows Xavier~\citep{glorot2010understanding}. We run 100 rounds of FL with $5\%$ clients sampled per round for all algorithms. All results are reported in Table~\ref{tb:femnist}.

\begin{figure*}[t]
\centering
\begin{adjustbox}{valign=t}
\begin{minipage}[t]{0.55\linewidth}
\caption{Test accuracy(\%)$\pm$ std}
\label{tb:cfl-results}
\resizebox{\linewidth}{!}{
\begin{tabular}{lccc}
\toprule[\heavyrulewidth]
    & \textbf{~Shifted MNIST~}     & \textbf{~Shifted CIFAR~}   \\
\cmidrule(r){2-4}
$N$, $|D|$    & 20, 9600          & 20, 8000& \\ 
\midrule
FedAvg  &   24.44$\pm$0.03  &   13.01$\pm$2.48  \\
{FedInit}  &   {25.13$\pm$0.02}  &   {13.53$\pm$1.24} \\
IFCA(4)    &    98.11$\pm$0.03   &  54.05$\pm$0.59  \\
{FedSEM(4)}    &    {98.03$\pm$0.05}   &  {54.79$\pm$0.97}  \\
CFL     &   98.20$\pm$0.02  &  55.23$\pm$1.02     \\
StoCFL  &   98.14$\pm$0.01 &  55.42$\pm$1.42   \\ 
\bottomrule[\heavyrulewidth]
\end{tabular}}
\end{minipage}
\end{adjustbox}
\hspace{5pt}
\begin{adjustbox}{valign=t}
\begin{minipage}[t]{0.4\linewidth}
\centering
\includegraphics[width=\linewidth]{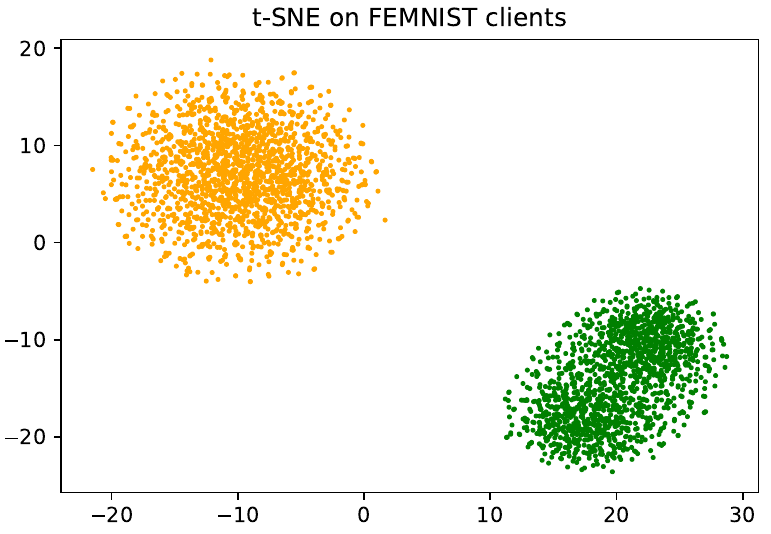}
\caption{Clustering Results of StoCFL on FEMNIST.}
\label{fig:femnist}
\end{minipage}
\end{adjustbox}
\end{figure*}

\textbf{Discussion on FEMNIST results.}  
We observed that StoCFL achieves better performance on FEMNIST. Moreover, CFL requires more resources as it requires the full participation of clients and IFCA achieves worse performance with an inaccurate $\tilde{M}$. Besides, StoCFL is better than FedAvg because it fully uses the data across clients via the global model, while CFL and IFCA are worse than FedAvg.
For the client clustering results, we observe that StoCFL clusters FEMNIST clients into two main clusters as shown in Figure \ref{fig:femnist}. It indicates that FEMNIST consists of two implicit data-generating distributions. Importantly, this finding aligns with previous studies~\citep{marfoq2021federated, ghosh2020efficient} on FEMNIST. 

{
\textbf{Discussion on FedCelebA results.} 
The FedCelebA task is a binary classification task, which is relatively easier than the FEMNIST task. Consequently, we observed a minor performance gap across all methods. Nevertheless, StoCFL still implements the highest testing accuracy of all methods. Moreover, we find that different SFCC parameters $\tau$ can lead to different numbers of client clusters in the FedCelebA task. For example, using a small $\tau$, SFCC outputs two main client clusters. While using a relatively large $\tau$, SFCC outputs four main client clusters. We further explain this phenomenon in the next section. Importantly, StoCFL performance is robust to the selection of $\tau$, as it leverages the global model to share knowledge across clusters.
}

In all, the real-world dataset evaluation results reveal the ability of StoCFL to handle non-synthetic datasets and further indicate the practical ability of StoCFL.

\begin{table}[t]
\caption{Test accuracy(\%)$\pm$ std on FEMNIST and CelebA tasks.}\label{tb:femnist}
\resizebox{\linewidth}{!}{
\begin{tabular}{ccccccccc}
\toprule
        & FedAvg & CFL & IFCA(2) & IFCA(3) & FedSEM(2) & FedSEM(3) & StoCFL($\tau=0.55$) & StoCFL($\tau=0.60$) \\
FEMNIST & 86.27$\pm$0.11 &  79.64$\pm$0.37 & 73.11$\pm$0.92   &   69.11$\pm$2.49   &   85.81$\pm$1.12    &  88.14$\pm$1.32  & 90.92$\pm$0.02 & 90.71$\pm$0.04     \\ 
FedCelebA  & 87.48$\pm$0.41   & 88.71$\pm$0.73 &  88.38$\pm$0.89      &   85.38$\pm$1.39         &   89.20$\pm$1.21           &    88.2$\pm$2.37         & 91.12$\pm$0.47           &  91.73$\pm$0.17   \\ \bottomrule
\end{tabular}}
\end{table}

\subsection{Impact of Hyper-parameters}\label{sec:hyper}

In this section, we first study the influence of regularization weight $\lambda$ on model performance. Then, we illustrate the impacts of clustering threshold $\tau$ on clustering results. Based on the results, we provide insights about the tuning of hyper-parameters when deploying StoCFL on applications.

\textbf{The $\lambda$ adjusts the impacts of the global model on cluster models}. We empirically demonstrate the effect of $\lambda$ via experiments on Pathological-MNIST, Rotated-MNIST, Shifted-MNIST, and Hybrid-MNIST settings (described in Section~\ref{sec:stoclu}). We run 50 communication rounds of StoCFL with $\lambda = \{0, 0.01, 0.05, 0.5, 1, 10\}$, and report the test accuracy of the global model and cluster models. 

The results are summarized in Table \ref{tab:effectl}. The global accuracy is the mean performance of the final $\boldsymbol{\omega}$ on all augmented test sets, while the contents of other columns are the average performance of cluster models. We note that $\lambda=0$ indicates the conventional CFL with correct client clustering results. In comparison, the results with $\lambda > 0$ prove that StoCFL improves the cluster model's performance by introducing useful knowledge from other clusters via the regularization term. For instance, the global accuracy results of Rotated-MNIST and Shifted-MNIST are distinct, while the cluster models are enhanced by the regularization term with $\lambda=0.05$. Besides, compared with the conventional CFL results ($\lambda=0$), the results of StoCFL are better. 

Furthermore, the value of $\lambda$ where the cluster models achieve the best accuracy is not the same as the impacts of decentralized data distributions in these four settings are different. Hence, we conclude that the best $\lambda$ relies on the real scenario of Non-IID data. In the real-world training process of StoCFL, the $\lambda$ could be adjusted dynamically during the training process. Besides, we could refer to ~\citep{li2020federated, li2021ditto} for the best strategies for choosing $\lambda$.

\begin{table}[h]
\centering
\caption{Effect of $\lambda$ to Test accuracy (\%).}\label{tab:effectl}
\resizebox{0.8\linewidth}{!}{
\begin{tabular}{lccccccc}
\toprule[\heavyrulewidth]
\multirow{2}{*}{\textbf{~MNIST~}} & \multirow{2}{*}{Global} & \multicolumn{5}{c}{StoCFL, $\lambda$}  &              \\
\cmidrule(r){3-8}
    &   & 10 & 1 & 0.5 & 0.05 & 0.01 & 0 \\ 
\midrule
Pathological       & 92.17  & \textbf{89.28}  & 86.89    & 83.49    & 37.83     & 24.12   &  24.05 \\
Rotated            & 92.65  & 62.06  & 94.88    & 95.28    & \textbf{95.86}     & 94.20   &  92.25 \\
Shifted            & 24.46  & 32.69  & 92.36    & 93.85    & \textbf{95.12}     & 93.80   &  92.22 \\ 
Hybrid             & 92.50  & 92.26  & 92.75    & \textbf{92.77}    & 92.69     & 91.99   &  90.11 \\ 
\bottomrule[\heavyrulewidth]
\end{tabular}}
\end{table}

\begin{figure}[htbp]
\centering
\includegraphics[scale=0.7]{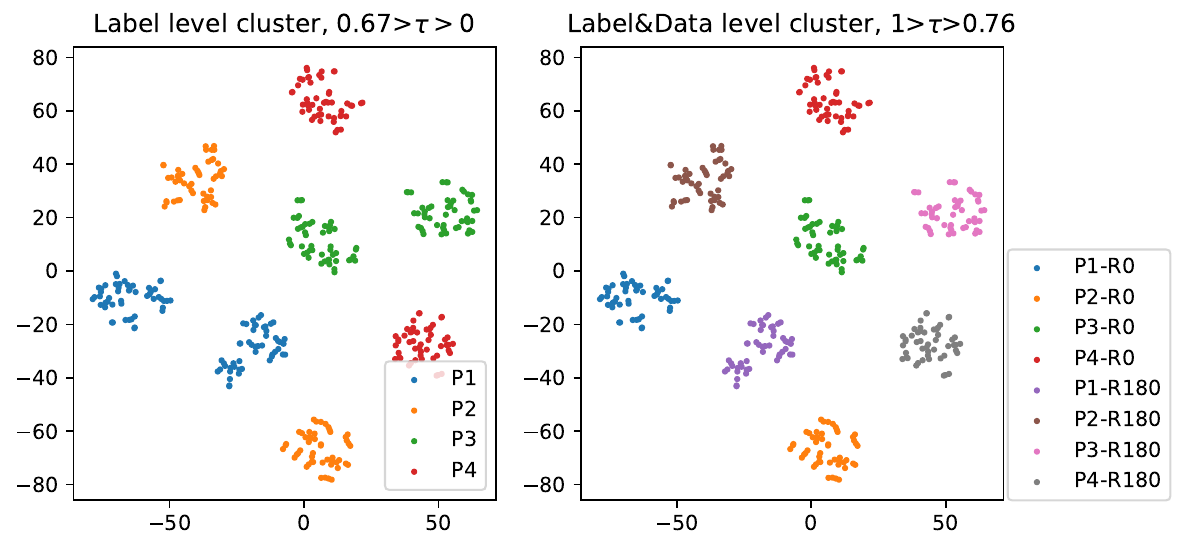}
\caption{The value of $\tau$ decides the clustering granularity.}
\label{fig:compare_with_cfl_gap}
\end{figure}

\textbf{The clustering threshold $\tau$ determines the clustering granularity.} We demonstrate the effects of $\tau$ on client clustering with a hybrid Non-IID setting combination. In this part, we follow the partition strategies described in Section~\ref{sec:stoclu}. Firstly, we create Rotate-MNIST by 0,180 degrees of rotation and marked them with R0, R180. Then, we pathologically partition (marked by P1, P2, P3, P4) each of them into 200 clients. In this case, we have 400 clients in total, where the client data distributions are 2 clusters in feature distribution (rotation degree), and 4 clusters in label distribution. From the overall perspective, there are 8 clusters in both label and feature distribution. We conduct the stochastic federated client clustering procedure with different $\tau$ on this setting. In our experiments, we observe that the clustering results vary with the value of $\tau$. Then, we depict the representative results via t-SNE in Figure~\ref{fig:compare_with_cfl_gap} and color the points according to the clustering results.

The results indicate that the value $\tau$ determines the focus of the clustering algorithm. For instance, $\tau > 0.76$ in this case, the clustering algorithm will cluster clients only when their feature distribution and label distribution are similar. In contrast, $\tau<0.67$ makes the clustering algorithm focus on the label distribution while ignoring the differences in feature level. Hence, a higher value of $\tau$ decides the client clustering in a more fine-grained way. More importantly, we emphasize that StoCFL is robust to the client clustering results as shown in Table~\ref{tb:femnist}. For a further relation between the clustering granularity and the threshold $\tau$, we will study it in future work.

\subsection{Cluster Inference and Generalization to Unseen Clients}~\label{sec:inference}

We further evaluate the ability of StoCFL to infer newly joined clients as discussed in Section~\ref{sec:discuss}. We demonstrate the inference ability of StoCFL by studying the generalization performance to unseen clients in the FEMNIST and {FedCelebA} setting.

\begin{table}[h]
\centering
\caption{Generalization performance (\%) on FEMNIST and FedCelebA}\label{tab:gener}
\resizebox{0.8\linewidth}{!}{
\begin{tabular}{llcccc}
\toprule
Setting & Metric                 & FedAvg & CFL & IFCA(2) & StoCFL \\ \hline
\multirow{2}{*}{FEMNIST} & Test Acc.  & 85.93$\pm$0.16 & 79.78$\pm$0.28 & 74.82$\pm$0.49 & \textbf{91.00$\pm$0.05} \\
& Unseen Acc.   & 86.70$\pm$0.22 & 73.24$\pm$0.99 & 75.78$\pm$0.45 & \textbf{91.06$\pm$0.05}\\
\multirow{2}{*}{{FedCelebA}} & Test Acc.  & {88.79$\pm$1.25} & {89.38$\pm$0.98} & {89.82$\pm$0.49} &  {\textbf{90.79$\pm$0.31}} \\
& Unseen Acc.   & {89.65$\pm$1.32} & {85.73$\pm$1.99} & {88.29$\pm$0.45} & {\textbf{91.46$\pm$1.09}}\\
\bottomrule
\end{tabular}}
\end{table}

In particular, we randomly selected 30\% clients as test clients who did not participate in the FL process. Only 70\% clients are participating in the CFL process. {We note that FedSEM does not support the cluster inference method for nonparticipant clients.} Hence, we only run StoCFL, CFL, and IFCA to infer the cluster of unparticipated clients using their training data samples.  As a result, we report in Table~\ref{tab:gener} the accuracy of test data samples from participated (marked by \textit{Test Acc.}) and unanticipated clients (marked by \textit{Unseen Acc.}). According to the results, we observed that FedAvg and IFCA implement consistent test accuracy and unseen accuracy. And, CFL achieves a lower generalization performance of unseen accuracy. Importantly, StoCFL achieves better generalization performance while preserving higher test accuracy.

\section{Future Works \& Conclusion}

\textbf{Future works.} 
We plan to improve the framework from optimization acceleration and system aspects. For instance, we can combine knowledge transfer techniques in local optimization for better cluster models. Besides, establishing a dynamic join-leave mechanism can further enhance StoCFL to deal with changing the local data samples. Furthermore, such a dynamic join-leave mechanism could exclude potential Byzantine clients from a benign cluster and obtain more robust FL~\citep{hoang2024effective}. {Furthermore, federated prompt-tuning~\citep{wengprobabilistic} presents a promising approach for tackling complex tasks with moderate data, particularly in the era of large pretrained models~\citep{ma2023federated}. Extending StoCFL to optimize heterogeneous prompts across FL clients for reprogramming pretrained models' behavior offers significant potential. Additionally, exploring client clustering methods like SFCC to mitigate the impact of non-IID data on pretrained models could provide valuable insights.}

In conclusion, we proposed a novel CFL framework, StoCFL, consisting of stochastic federated client clustering and bi-level CFL procedure. Our study demonstrates that StoCFL could cluster federated clients with different unknown distributions and train better-generalized models. Besides, StoCFL is flexible and robust with real-world applications, especially in a varying FL system. Furthermore, the results of intensive experiments have shown the superiority of the proposed algorithms.


\section*{Acknowledgements}
This work was partially supported by the Major Key Project of PCL (No. 2022ZD0115301), and
the National Key Research and Development Program of
China (No. 2018AAA0100204).

\bibliographystyle{plainnat}
\bibliography{references}
\end{document}